\documentclass[a4paper,11pt]{article}
\usepackage[utf8]{inputenc} 
\usepackage[T1]{fontenc}    
\usepackage[margin=1in]{geometry}
\usepackage{wrapfig}
\usepackage{authblk}
\usepackage{import}
\usepackage{microtype}
\usepackage{graphicx}
\usepackage{xcolor}
\usepackage{booktabs} 
\usepackage[algo2e]{algorithm2e}
\usepackage{hyperref}
\usepackage{comment}
\usepackage{subfig}
\usepackage{enumitem}
\setlist{leftmargin=3mm}
\usepackage{soul}

\usepackage{caption}
\usepackage{multirow}

\usepackage{amsmath,amsthm,amssymb,amsfonts}
\usepackage{algorithm}
\usepackage{algorithmic}

\usepackage{comment}
\usepackage{bm}
\usepackage{pifont}

\theoremstyle{plain}
\newtheorem{theorem}{Theorem}[]

\newtheorem{definition}{Definition}[]

\newtheorem{fact}{Fact}[]

\def\1{\mathbf{1}}

\def\vb{{\mathbf{b}}}
\def\vc{{\mathbf{c}}}

\def\ve{{\mathbf{e}}}

\def\vm{{\mathbf{m}}}

\def\vv{{\mathbf{v}}}
\def\vw{{\mathbf{w}}}

\def\mC{{\mathbf{C}}}

\def\mI{{\mathbf{I}}}

\def\mV{{\mathbf{V}}}
\def\mW{{\mathbf{W}}}

\DeclareMathAlphabet{\mathsfit}{\encodingdefault}{\sfdefault}{m}{sl}
\SetMathAlphabet{\mathsfit}{bold}{\encodingdefault}{\sfdefault}{bx}{n}

\def\sA{{\mathbb{A}}}
\def\sB{{\mathbb{B}}}

\def\sR{{\mathbb{R}}}
\def\sS{{\mathbb{S}}}

\newcommand{\E}{\mathbb{E}}

\DeclareMathOperator*{\argmax}{arg\,max}
\DeclareMathOperator*{\argmin}{arg\,min}

\makeatletter

\newcommand{\printfnsymbol}[1]{%
  \textsuperscript{\@fnsymbol{#1}}%
}
\makeatother
\begin{document}

\title{DISCO : efficient unsupervised decoding for \underline{dis}crete natural language problems via \underline{co}nvex relaxation}
\date{}
\author[1]{Anish Acharya\thanks{equal contribution}}
\author[1]{Rudrajit Das\printfnsymbol{1}}
\affil[1]{University of Texas at Austin}
\maketitle
\begin{abstract}
In this paper we study test time \textit{decoding}; an ubiquitous step in almost all sequential text generation task spanning across a wide array of natural lamguage processing (\textsc{NLP}) problems.
Our main contribution is to 
develop a continuous relaxation framework for the combinatorial \textsc{NP}-hard decoding problem and propose \textsc{Disco} - an efficient algorithm based on standard first order gradient methods. 
We provide tight analysis and show that our proposed algorithm linearly converges to within $\epsilon$ neighborhood of the optima. Finally, we perform preliminary experiments on the task of adversarial text generation and show superior performance of \textsc{Disco} over several popular decoding approaches.

\end{abstract}

\section{Introduction}
Consider the following setup: We are given a ground set $\E$ and a \textit{black box oracle} set function $f: 2^{\E} \to \sR$ and some constraints $\mathcal{I} \subseteq 2^{\E}$ assumed to be down monotone. 
Then \textit{constrained set function maximization} \cite{schrijver2003combinatorial} refers to the following optimization problem:
\begin{equation}
    \label{equation:set_max}
    \max f(\sS) \; \; \text{s.t.} \; \sS \in \mathcal{I}
\end{equation}
We claim that the decoding problems arising in several common \textsc{NLP} applications can be directly mapped to the \eqref{equation:set_max}. 

To see this, first consider a simple combinatorial \textbf{text generation} approach - a classic use case for sequential decoding \cite{lei2018discrete} : The input is a piece of text $\sS = \{e_1, e_2 , \dots , e_n\} \in \mathcal{X}^n$ composed of $n$ discrete features where feature space $\mathcal{X}$ denotes the vocabulary. For example,  Suppose, for each feature $e_i \in \mathcal{X}$ we have $k_i$ possible replacement candidates (ex. synonyms) denoted by $e_i^j, \; \forall j \in [k_i]$ with $e_i^0 = e_i, \; \forall e_i \in \mathcal{X}$. Let,
$T$ denote a valid transformation of $\sS$ defined as a combined replacement of each individual feature $e_i; \; i\in [n]$. Let, $T$ denote a valid transformation of $\sS$ defined as a combined replacement of each individual feature $e_i; \; i\in [n]$. Note that, we can index $T$ by a vector $\mI \in \prod_{i=1}^n\{0, 1, ... , k_i\}$ list of all possible such combinations of valid replacements. 

Further, it is assumed that we are given a black box oracle $f$ that returns a score for each input transformation. Now, we want to solve the following maximization problem:
\begin{equation}
\label{eq:adv_attack}
    \argmax_{\|\mI\|_0 \leq m}f(T_{\mI}(\sS))
\end{equation}

As a concrete application, consider the adversarial text generation problem where $\sS$ is a piece of text (ex. sentence) composed of atoms (ex. words) and each atom has $k-1$ replacement candidates (ex. synonyms). $f$ is the model (ex. sentence classifier) i.e. $f(\sS)$ denote the score (ex. loss) and the goal is to perturb at most $m$ atoms to obtain a transformed sentence $T_{\mI}(\sS)$ such that the loss is maximized i.e. would flip the confidence (ex. make the classifier predict wrong label) of the model and as a result $T_{\mI}(\sS)$ would be an adversarial example of $\sS$.

\label{sec:intro}

\section{Preliminaries}
In this section we introduce some notations and recall some concepts that will be used throughout the manuscript. 
\begin{definition}
\label{def:marginal_gain}
    \textbf{\textit{Discrete Derivative.}} For a set function $f: 2^{\E} \to \sR \;, \sS \subseteq \E \;, e \in \E\setminus\sS$ the discreet derivative or the marginal gain of $f$ at $\sS$ with respect to $e$ is defined as:
    \begin{equation}
        \label{marginal_gain}
        \Delta_f(e | \sS) = f(\sS \cup \{e\}) - f(\sS) 
    \end{equation}
\end{definition}

\begin{definition}
\label{def:sub_f}
    \textbf{\textit{Submodularity.}} A set function $f: 2^{\E} \to \sR \;$ is submodular if $\forall \sA \subseteq \sB \subseteq \E \;, e \in \E\setminus\sB$ the following holds:
    \begin{align}
        &\Delta_f(e | \sA) \geq \Delta_f(e | \sB)
    \end{align}
    It can be shown that it is equivalent to the following condition:
    \begin{align}
        f(\sA) + f(\sB) \geq f(\sA \cap \sB) + f(\sA \cup \sB)
    \end{align}
\end{definition}

\begin{fact}
\label{def:lin_comb_sm}
submodularity is preserved under taking non-negative linear combinations i.e. If $f_1, f_2, \dots , f_n$ are submodular then linear combination $f = \sum_{i=1}^n \alpha_i f_i$ is also submodular $\forall \alpha_i \geq 0$
\end{fact}

\begin{definition}
    \textbf{\textit{Embedding Layer.}}  Learning dense vector representation of atoms is the key in the recent success of Natural Language processing. The idea is quiet simple, given an input i.e. one hot encoding over vocabulary, embedding is a mapping to a dense continuous vector space $V: \mathcal{X}^n \to \sR^d$. 
    Early work include using ideas like SVD to learn compact representation of highly sparse one-hot input space (See \cite{bengio2003neural} and references therein). However, post Deep Learning era, Neural Networks trained with context to learn word embedding has proven to be a breakthrough in NLP research and now one usually can just use pre-trained word embedding maps to map discrete input space to embedding space using Word2Vec style embedding models \cite{mikolov2013efficient, mikolov2013distributed}. 
\end{definition}

\paragraph{Set Function Maximization}
Consider the following setup: We are given a ground set $\E$ and a black box oracle set function $f: 2^{\E} \to \sR$. Also suppose we are given constraints $\mathcal{I} \subseteq 2^{\E}$ assumed to be down monotone. Then in set function maximization one wish to solve the following problem:
\begin{equation}
\label{eq:set_max}
    \begin{aligned}
    &\max_{\sS \subseteq \mathcal{I}} f(\sS) \quad \text{s.t.} \; &\sS \in \mathcal{I}
    \end{aligned}
\end{equation}

For instance, the simplest constraint can be the uniform matroid constraint i.e. the cardinality constraint $|\sS| \leq k$ where $k \leq |\E|$. 

\paragraph{Submodular Maximization}
In submodular maximization one wishes to solve \eqref{eq:set_max} when $f$ is a submodular set function admitting to definition \ref{def:sub_f}. For many class of submodular functions even this simplest problem has been shown to be NP hard \cite{krause2014submodular, krause2012near, feige1998threshold}. Thus, often the goal of research in the area is to find efficient approximate algorithms to solve \eqref{eq:set_max}. 

\begin{algorithm}[H]
    \SetAlgoLined
    \textbf{input:} $\E, \; \mathcal{I}\subseteq2^{\E}, \;f:2^{\E} \to \sR$ \\
    $\sA \gets \emptyset, \; \sS \gets \emptyset$ \\
    $\sA \gets \{e : \sS \cup e \in \mathcal{I}\}$\\
    \While{$\sA \neq \emptyset$}
    {
        $\sS = \sS \cup \{\argmax_{e} \Delta(e | \sS)\}$
    }
    \caption{Greedy Set Function Maximization}
    \label{algo:greedy}
\end{algorithm}\DecMargin{1em}

The popular Greedy algorithm to solve \eqref{eq:set_max} approximately (Algorithm \ref{algo:greedy}), starts with an empty set $\sS = \emptyset$ and at iteration $t$ add the element that maximizes $\Delta(e | \sS_{t-1})$ i.e. given by the following update equation:
\begin{equation}
\label{eq:greedy_iter}
\begin{aligned}
\end{aligned}
    \sS_t = \sS_{t-1} \cup \{\argmax_{e} \Delta(e | \sS_{t-1})\}
\end{equation} 

The celebrated result \cite{nemhauser1978best} established the approximation guarantee of greedy for submodular maximization.

\begin{theorem}
\label{theo:greedy_monotone}
\cite{nemhauser1978best}
Given a nonnegative monotone submodular function $f: 2^{\E} \to \sR_{+}$ and $\{\sS_i\}_{i \geq 0}$ be greedily selected sets defined by the iterations \eqref{eq:greedy_iter}:
\begin{equation}
\label{eq:greedy_classic}
    f(\sS_k) \geq (1 - \frac{1}{e})\max_{\sS : |\sS| \leq k} f(\sS)
\end{equation}
\cite{krause2014submodular} extends this result for any iteration t 
\begin{equation}
    \label{eq:greedy_classic_general}
    f(\sS_t) \geq (1 - e^{- \frac{t}{k}})\max_{\sS : |\sS| \leq k} f(\sS)
\end{equation}
\end{theorem}

\begin{proof}
For the proof we refer the reader to \cite{krause2014submodular}. 
\end{proof}

Going beyond uniform matroid constraint, suppose $(E, \mathcal{I}_1), (E, \mathcal{I}_2, \dots , (E, \mathcal{I}_p)$ are p matroids with $\mathcal{I} = \cap_i \mathcal{I}_i$. Even though now $(E, \mathcal{I}$ is no longer a matroid in general, it can be shown that greedy guarantees a solution within $\frac{1}{p+1}$ factor of the optima. i.e. $f(\sS_G) \geq \frac{1}{p+1} \max_{\sS\in \mathcal{I}}f(\sS)$. This result also holds when $(E, \mathcal{I}$ is a p-extensible system \cite{calinescu2007maximizing}.

\section{Submodular Formulation}
We first note that problem \eqref{eq:adv_attack} can equivalently be formulated as: 
\begin{equation}
    \label{eq:adv_attack_qi}
    \argmax_{|X| \leq m} f(X)
\end{equation}
where we can define the set function $f: 2^{[n]} \to \sR, \; f(X) = \max_{supp(\mI)\subset X}f(T_{\mI}(\sS))$. The set function represents the model output for the input set of transformations $X$. Thus, we reduced the problem to searching over all possible sets of transformations of up to $m$ replacement to maximize the loss. \\
Note that for NLP models the first layer is usually the embedding layer that converts the input into word embedding space thus \eqref{eq:adv_attack_qi} for most neural networks is equivalent to:
\begin{equation}
    \label{eq:adv_attack_qi_embed}
    f(X) = \max_{supp(\mI)\subset \sS}f(\mV(T_{\mI}(\sS)))
\end{equation}

\begin{theorem}
\cite{lei2018discrete} For a general model $f$ , the problem \eqref{eq:adv_attack} can be polynomially reduced from subset sum and thus NP-hard. 
\end{theorem}
\begin{proof}
We take a very simple function $f$ and show that it could be reduced from the subset sum problem when $k \geq 2$. As an example let us take the simplest setup where given (a sentence) $\sS$ the model generates an (sentence) embedding vector as the average of individual token (word) embedding and the goal is to maximize the $\ell_2$ distance from a target embedding \cite{iyyer2015deep}:    
\begin{equation}
    f(X) = \argmax_{\text{supp}(\mI)\subset \sS} \left\| \sum_{i=1}^n \vv_i^j -\vv \right\|^2
\end{equation}
Suppose, there is an algorithm that solves this problem in polynomial time.  Let the $n$ indexes in the transformation be $s_1, s_2, \dots , s_n$ and the target be $\mW$. We can let $\vv_i^0 = [s_i, 0 , \dots, 0]$ and $\vv_i^j = [0, \dots, 0]; \; j= 1, \dots , k-1$ and target $\vv = [\mW , 0, \dots, 0]$. Now it is immediate that the best approximation of $\vv$ is exactly $\vv$ will suffice the subset sum problem. Thus it contradicts with the fact that subset sum is NP complete.
\end{proof}

\begin{theorem}
\label{theo:qi_greedy} If $f$ in \eqref{eq:adv_attack} is submodular monotone then greedy would achieve a $(1 - \frac{1}{e})$ approximation in polynomial time.
\end{theorem}
\begin{proof}
The proof is straightforward the moment we re-formulated the problem \eqref{eq:adv_attack} to the problem \eqref{eq:adv_attack_qi}. This is maximizing a set function over a uniform matroid constraint. Thus, if $f$ is monotone submodular then by using Theorem \ref{theo:greedy_monotone}, we conclude that greedy achieves at least $(1 - \frac{1}{e})$ approximation of the optimal in polynomial time. 
\end{proof}

\subsection{Submodular Neural Networks}
Recently, there has been several efforts to study submodular properties of Deep Networks.~\cite{bilmes2017deep, dolhansky2016deep} establish a class of feed forward networks that can be approximated as submodular functions. Specifically~\cite{lei2018discrete} show submodular properties of two neural networks that are of particular interest in practical applications. 
\begin{fact}
\textup{\cite{lei2018discrete}}
Consider the following simple version of W-CNN \textup{\cite{kim2014convolutional}}: let the stride be $s$, window size $h$ and the word embedding of $i-$th word be $\vv_i$. Then output of convolution layer is given as $\mC = [c_{ij}]_{i \in [n/s], j\in [m]}$ from $n$ words and $m$ filters. 
\begin{equation*}
    c_{ij} = \phi(\vw_j^T \vv_{s(i-1)+1:s(i-1)+h} + b_j), \; i = 1,2,\dots,n/s
\end{equation*}
$\vw_j \in \sR^{Dh}$ is the $j$-th filter, $b_j$ is bias term and $\phi$ is non-linear activation. $\vv_{i:j}$ denotes the concatenation of word vectors in the
window of words $i$ through $j$. Each filter $\vw_j$ is applied to individual
windows of words to produce feature map $\vc^j = [c_{1j}, c_{2j}, \dots, c_{(n/s)j}]^T$. This feature map is applied to a max pooling layer to obtain:
\begin{equation*}
    \hat{c_j} = \max_i c_{ij} 
\end{equation*}
Thus the W-CNN without dropout and softmax is given as: 
\begin{equation*}
    f(\vv_{1:n}) = \vw^T \cdot \hat{\vc} + \vb^T
\end{equation*}
Now, suppose there’s no
overlapping between each window and we only look at
transformations that will increase the output then: 
$f(\sS) = \max_{supp(\mI)\subset \sS}f^{WCNN}(\mV(T_{\mI}(\sS)))$ is submodular. 
\end{fact}
\begin{proof}
The proof idea is that every coordinate in $\hat{\vc}$ is a combination of max pooling over modular functions and hence submodular. Next, the output is sum of submodular functions then using Definition \ref{def:lin_comb_sm} it is also a submodular function. For a complete proof we refer the reader to Section A.4. of \cite{lei2018discrete}. 
\end{proof}

\begin{fact}
\textup{\cite{lei2018discrete}} Consider a RNN with $T$ time steps and hidden layer of single node. Then for al $t \leq T$, given the previous hidden state $h_{t-1}$ and input embedding $\vv_{t-1}$ RNN updates the hidden state as:
\begin{equation*}
    h_t = \phi(wh_{t-1} + \vm^T \vv_{t-1} + b)
\end{equation*}
Then the output of RNN is characterized as 
\begin{equation*}
    f(\vv_{1:T}) = y h_T
\end{equation*}
if $w, y$ are positive and the activation is a non-decreasing concave function then \\
$f(\sS) = \max_{supp(\mI)\subset \sS}f^{RNN}(\mV(T_{\mI}(\sS)))$ is submodular.
\end{fact}
\begin{proof}
The proof proceeds as follows: First, it can be shown 
that the same amount of change induced on
an intermediate layer has a diminishing return property when the network is perturbed at other features. Then together
with the concavity and non-decreasing property of the network one can finish the proof. For a complete proof we refer the reader to Section A.5. of \cite{lei2018discrete}.
\end{proof}

\label{sec:submodular_qi}
\section{DISCO: Convex Relaxation}
As discussed in Section \ref{sec:submodular_qi}, the apparent computationally intractable problem \eqref{eq:adv_attack} can be efficiently solved using Greedy Algorithm \ref{algo:greedy} with $(1 - \frac{1}{e})$ approximation to the optimal whenever $f$ is submodular. It was also shown that for some class of simplified Neural Networks one can show some submodular properties. That being said, so far the known submodular approximations of real work networks is limited and there is no clear theory to indicate if all the neural networks can be reasonably approximated by submodular functions. With this observation we propose to relax this discrete combinatorial optimization problem into a approximately continuous problem and claim that it can be solved using a stochastic first order continuous optimization method. 

\subsection{Formulation using Extension Framework}
First recall  how we can use the concept of extensions on set functions $f(\sS): 2^{\E} \to \sR$. By identifying sets $\sS$ with binary vectors $\ve_{\sS}$ in which $i-$th component is 1 if $i \in \sS$ and 0 otherwise, now we can equivalently represent $f$ as a function defined over corners of the unit cube $\Tilde{f}: \{0, 1\}^n \to \sR$ where $n = |V|$ and $\Tilde{f}(\ve_{\sS}) = f(\sS)$. Now, this can be extended to the entire unit cube $[0, 1]^n$. 

Leveraging this idea, we re-formulate \eqref{eq:adv_attack} follows: 
\begin{align}
    \label{eq-2}
    & \max_{\alpha_{i}^{(j)} \text{ } \forall \text{ }i,j} f(\sum_{j=1}^{k_{1}}\alpha_{1}^{(j)}\bm{w}_{1}^{(j)},\ldots,\sum_{j=1}^{k_{n}}\alpha_{n}^{(j)}\bm{w}_{n}^{(j)})
    \\
    \label{eq-3}
    \text{ s.t.} \quad &\alpha_{i}^{(j)} = \{0,1\} \text{ }\forall \text{ }i \in [n], \text{ }j \in [k_{i}]
    \\
    \label{eq-4}
    & \sum_{j=1}^{k_{i}}\alpha_{i}^{(j)} = 1 \text{ }\forall \text{ }i \in [n]
\end{align}
Note that this relaxation is same in spirit of Lova\'{s}z extension \cite{lovasz1983submodular}, multilinear extension \cite{vondrak2008optimal}. \\
Now, we can relax the binary valued $\alpha_{i}^{(j)}$ to $\widetilde{\alpha}_{i}^{(j)} \; \forall \text{ }i \in [n],\text{ }j \in [k_{i}]$ parameterized as:
\begin{equation}
    \label{eq-5}
    \widetilde{\alpha}_{i}^{(j)} = \frac{{\big(\beta_{i}^{(j)}\big)}^{2p}}{Z_{i}}, Z_{i} = \sum_{j=1}^{k_{i}} {\big(\beta_{i}^{(j)}\big)}^{2p}
\end{equation}
In (\ref{eq-5}), $p \in \mathbb{N}$ is a hyper-parameter of our choice. Observe that the $\widetilde{\alpha}_{i}^{(j)}$ obey (\ref{eq-4}). Then, the relaxed objective function is as follows:
\begin{align}
    &\argmax_{{\beta}_{i}^{(j)}\; \forall \text{ }i,j}\left[ f(\sum_{j=1}^{k_{1}}\widetilde{\alpha}_{1}^{(j)}\bm{w}_{1}^{(j)},\ldots,\sum_{j=1}^{k_{n}}\widetilde{\alpha}_{n}^{(j)}\bm{w}_{n}^{(j)}) + \lambda \sum_{i=1}^{n}\sum_{j=1}^{k_{i}}|\beta_{i}^{(j)}|\right]
    \\
    \nonumber
    & \text{ where } \widetilde{\alpha}_{i}^{(j)} = \frac{{\big(\beta_{i}^{(j)}\big)}^{2p}}{Z_{i}}, Z_{i} = \sum_{j=1}^{k_{i}} {\big(\beta_{i}^{(j)}\big)}^{2p}.
\end{align}
Also note that we can instead define $C(x) = -f(x)$ and instead solve the following problem:
\begin{align}
\label{eq-6}
    &\argmin_{{\beta}_{i}^{(j)}\; \forall \text{ }i,j}\left[ C(\sum_{j=1}^{k_{1}}\widetilde{\alpha}_{1}^{(j)}\bm{w}_{1}^{(j)},\ldots,\sum_{j=1}^{k_{n}}\widetilde{\alpha}_{n}^{(j)}\bm{w}_{n}^{(j)}) + \lambda \sum_{i=1}^{n}\sum_{j=1}^{k_{i}}|\beta_{i}^{(j)}|\right]
    \\
    \nonumber
    & \text{ where } \widetilde{\alpha}_{i}^{(j)} = \frac{{\big(\beta_{i}^{(j)}\big)}^{2p}}{Z_{i}}, Z_{i} = \sum_{j=1}^{k_{i}} {\big(\beta_{i}^{(j)}\big)}^{2p}.
\end{align}
\subsection{Discrete Output Space}
The solution of (\ref{eq-6}) is not guaranteed to be one-hot like we want. In other words, relating to the previously stated discrete version of the problem, 
we want $\alpha_{i}^{(j)} \in \{0,1\}$ but ${\big(\beta_{i}^{(j)}\big)}^{2p}/Z_{i}$ is not going to be either 0 or 1 almost surely. To handle this, we greedily do the following:
\begin{equation}
\label{eq-7}
     j_{i} = \argmax_{j \in [k_{i}]} |{\beta_{i}^{(j)}}| \text{ and }
    \alpha_{i}^{(j)} = \delta(j - j_{i}) \text{ } \forall \text{ } i \in [n]. 
\end{equation}
\subsection{Theoretical Guarantees}
\label{sec:theory}
We shall introduce some definitions before we present our main result.
\small
\begin{align}
    \label{eq-8}
    & f(\bm{\beta}) = C(\sum_{j=1}^{k_{1}}\widetilde{\alpha}_{1}^{(j)}\bm{w}_{1}^{(j)},\ldots,\sum_{j=1}^{k_{n}}\widetilde{\alpha}_{n}^{(j)}\bm{w}_{n}^{(j)})
    \\
    \nonumber
    & \text{where } \widetilde{\alpha}_{i}^{(j)} = \frac{{\big(\beta_{i}^{(j)}\big)}^{2p}}{Z_{i}}, Z_{i} = \sum_{j=1}^{k_{i}} {\big(\beta_{i}^{(j)}\big)}^{2p} \text{ and } \bm{\beta} \in \mathbb{R}^{\sum_{i=1}^{n}k_{i}}
    \\
    \nonumber
    & \text{is the vector formed by stacking the } \beta_{i}^{(j)}\text{'s} \text{ } \forall \text{ } j \in [k_{i}], i \in [n].
\end{align}
\normalsize
\begin{equation}
    \label{eq-9}
    h(\bm{\beta}) = \lambda \sum_{i=1}^{n}\sum_{j=1}^{k_{i}}|\beta_{i}^{(j)}| = \lambda \|\bm{\beta}\|_{1}
\end{equation}
\begin{equation}
    \label{eq-10}
    \phi(\bm{\beta}) \triangleq{} f(\bm{\beta}) + h(\bm{\beta})
\end{equation}

In (\ref{eq-6}), we are looking to minimize $\phi(\bm{\beta})$ without any constraints on $\bm{\beta}$. Notice that $\phi(c\bm{\beta}) < \phi(\bm{\beta})$ $\forall$ $0<c<1$. This is because with $c \in (0,1)$, $f(c\bm{\beta}) = f(\bm{\beta})$ and $h(c\bm{\beta}) < h(\bm{\beta})$. Due to this reason, we must impose some restrictions on $\bm{\beta}$ to have a minimizer for $\phi(\bm{\beta})$, at least for theory. However in practice, one may not impose any restrictions on $\bm{\beta}$ as we only care about the value of $f(\bm{\beta})$. 
The constraints that we impose on $\bm{\beta}$ to derive theoretical results are as follows:
\begin{equation}
    \label{eq-11}
    Z_{i} = \sum_{j=1}^{k_{i}} {\big(\beta_{i}^{(j)}\big)}^{2p} = 1 \text{ } \forall \text{ } i \in [n].
\end{equation}
Notice that the constraints defined in (\ref{eq-11}) above are convex. Let us now define the set of all feasible $\bm{\beta}$ obeying the constraints in (\ref{eq-11}) as:
\begin{equation}
    \label{eq-12}
    \mathcal{S} \triangleq{} \Big\{\bm{\beta}|Z_{i} = \sum_{j=1}^{k_{i}} {\big(\beta_{i}^{(j)}\big)}^{2p} = 1 \text{ } \forall \text{ } i \in [n]\Big\}.
\end{equation}
So finally, we wish to solve:
\begin{equation}
    \label{eq-13}
    \min_{\bm{\beta} \in \mathcal{S}} \text{ }\phi(\bm{\beta})
\end{equation}
We can solve (\ref{eq-13}) by using Algorithm \ref{alg:ppgd} inspired by projected proximal gradient descent . 

\begin{algorithm}[H]
  \caption{\textbf{DISCO - Proposed Projected Proximal Approach}}
  \begin{algorithmic}
    \STATE \textbf{Input}: number of iterations $T$, learning rate schedule $\eta_{k}$ for $k = 1,\ldots, T$, initial value of $\bm{\beta} = \bm{\beta}_{1}$.
    \vspace{0.2cm}
      \FOR{$k=1:T$}
        \vspace{0.2cm}
        \STATE
        {Compute $\nabla f(\bm{\beta_{k}})$.} 
        \vspace{0.3cm}
        \STATE
        {Update $\bm{\beta}_{k+1} = \text{P}_{\mathcal{S}}\{\text{prox}_{\eta_{k},h}(\bm{\beta}_{k} - \eta_{k}\nabla f(\bm{\beta_{k}}))\}$,}
        \STATE
        {where $\text{prox}_{\eta,h}(\bm{\beta}) = \text{argmin}_{\bm{\beta}'} \text{ } h(\bm{\beta}') + \frac{1}{2\eta}\|\bm{\beta} - \bm{\beta}'\|^{2}$,}
        \STATE
        {$\text{P}_{\mathcal{S}}(\bm{x}) = {\bm{x}}/{\|\bm{x}\|_{2p}}$ (i.e. the projection of $\bm{x}$ onto $\mathcal{S}$).}
        \vspace{0.2cm}
    \ENDFOR
  \end{algorithmic}
\label{alg:ppgd}
\end{algorithm}
Essentially, it follows the proximal gradient method followed by a projection step onto $\mathcal{S}$ (which is a very cheap operation) so that $\bm{\beta}_{k+1} \in \mathcal{S}$. Also, since $h(\bm{\beta}) = \lambda \|\bm{\beta}\|_{1}$, the $\text{prox}$ operator has a closed form expression which is as follows:
\begin{equation}
    \label{eq-14}
  \{\text{prox}_{\eta,h}(\bm{\beta})\}_{j}=\begin{cases}
    \{\bm{\beta}\}_{j} - \eta \lambda & \text{if $\{\bm{\beta}\}_{j} > \eta \lambda$}\\
    \{\bm{\beta}\}_{j} + \eta \lambda & \text{if $\{\bm{\beta}\}_{j} < -\eta \lambda$}\\
    0 & \text{otherwise}.
  \end{cases}
\end{equation}
In (\ref{eq-14}), $\{\text{prox}_{\eta,h}(\bm{\beta})\}_{j}$ and $\{\bm{\beta}\}_{j}$ are the $j^{\text{th}}$ elements of $\text{prox}_{\eta,h}(\bm{\beta})$ and $\bm{\beta}$, respectively.
We are now ready to present our main result which is as follows:

\begin{theorem}
Suppose $f(\bm{\beta})$ is convex and $L$-smooth (with respect to $\bm{\beta}$) over $\mathcal{S}$. Let 
$\bm{\beta}^{*} = \text{argmin}_{\bm{\beta} \in \mathcal{S}} \text{ }\phi(\bm{\beta})$.
Then Algorithm \ref{alg:ppgd} with $\bm{\beta}_{1} = \left[\frac{1}{k_{1}^{1/2p}}\mathbf{1}_{k_{1}},\frac{1}{k_{2}^{1/2p}}\mathbf{1}_{k_{2}},\ldots,\frac{1}{k_{n}^{1/2p}}\mathbf{1}_{k_{n}}\right]^{T}$ where $\mathbf{1}_{m}$ is the $m$-dimensional vector of all ones and constant step-size value $\eta < 1/L$ has the following convergence guarantee after $(T+1)$ iterations:
\[\phi(\bm{\beta}_{T+1}) - \phi(\bm{\beta}^{*}) \leq \frac{1}{2 \eta T} \sum_{i=1}^{n} k_{i}\Bigg(1+\frac{1}{k_{i}^{1/2p}}\Bigg)^{2}\]
Therefore, in $\mathcal{O}({\sum_{i=1}^{n}k_{i}}/{\epsilon})$ iterations, we have $\phi(\bm{\beta}_{T}) - \phi(\bm{\beta}^{*}) < \epsilon$.

This theorem illustrates that Algorithm \ref{alg:ppgd} requires $\mathcal{O}(\sum_{i=1}^{n}k_{i}/\epsilon)$ iterations to reach $\epsilon$-close to the optimal solution for the set function maximization formulation without any assumption on submodularity.
\end{theorem}
\begin{proof}
The proof is similar to that of convergence of proximal gradient descent.
Since $f$ is $L$-smooth over $\mathcal{S}$, we have for $\bm{\beta}_{k+1}$ and $\bm{\beta}_{k}$ (both $\in \mathcal{S}$):
\small
\begin{equation}
    \label{eq-15}
    f(\bm{\beta}_{k+1}) \leq f(\bm{\beta}_{k}) + \langle \nabla f(\bm{\beta}_{k}),\bm{\beta}_{k+1}-\bm{\beta}_{k}\rangle + \frac{L}{2}\|\bm{\beta}_{k+1}-\bm{\beta}_{k}\|^{2}
\end{equation}
\normalsize
Now since $f$ is also convex, we have for all $\bm{z} \in \mathcal{S}$:
\begin{equation}
    \label{eq-16}
    f(\bm{\beta}_{k}) \leq f(\bm{z}) - \langle \nabla f(\bm{\beta}_{k}),\bm{z}-\bm{\beta}_{k}\rangle
\end{equation}
Substituting this in (\ref{eq-15}), we get:
\small
\begin{equation}
    \label{eq-17}
    f(\bm{\beta}_{k+1}) \leq f(\bm{z}) + \langle \nabla f(\bm{\beta}_{k}),\bm{\beta}_{k+1}-\bm{z}\rangle + \frac{L}{2}\|\bm{\beta}_{k+1}-\bm{\beta}_{k}\|^{2}
\end{equation}
\normalsize
Let us define $\widehat{\bm{\beta}}_{k+1}$ as follows (recall that we are using constant step-size $\eta$):
\begin{equation}
    \label{eq-18}
    \widehat{\bm{\beta}}_{k+1} \triangleq{} \text{prox}_{\eta,h}(\bm{\beta}_{k} - \eta\nabla f(\bm{\beta_{k}}))
\end{equation}
Then, we have:
\begin{equation}
    \label{eq-19}
    {\bm{\beta}}_{k+1} = \text{P}_\mathcal{S}(\widehat{\bm{\beta}}_{k+1}) = \widehat{\bm{\beta}}_{k+1}/\|\widehat{\bm{\beta}}_{k+1}\|_{2p}
\end{equation}
Using the convexity of $h(\bm{\beta}) = \lambda \|\bm{\beta}\|_{1}$ in $\mathbb{R}^{\sum_{i=1}^{n}k_{i}}$, we have for $\bm{g} \in \partial h(\widehat{\bm{\beta}}_{k+1})$ and $\bm{z}$ as used in (\ref{eq-16}):
\begin{equation}
    \label{eq-20}
    h(\widehat{\bm{\beta}}_{k+1}) \leq h(\|\widehat{\bm{\beta}}_{k+1}\|_{2p}\bm{z}) + \langle \bm{g}, \widehat{\bm{\beta}}_{k+1} - \|\widehat{\bm{\beta}}_{k+1}\|_{2p}\bm{z}\rangle
\end{equation}
Now dividing (\ref{eq-20}) on both sides by $\|\widehat{\bm{\beta}}_{k+1}\|_{2p}$ and using the fact that $h(c \bm{\beta}) = c h(\bm{\beta})$ $\forall$ $c > 0$ along with (\ref{eq-19}), we get:
\begin{equation}
    \label{eq-21}
    h({\bm{\beta}}_{k+1}) \leq h(\bm{z}) + \langle \bm{g}, {\bm{\beta}}_{k+1} - \bm{z}\rangle
\end{equation}
Now adding (\ref{eq-21}) and (\ref{eq-17}), we get:
\begin{multline}
    \label{eq-22}
    \phi({\bm{\beta}}_{k+1}) \leq \phi(\bm{z}) + \langle \nabla f(\bm{\beta}_{k}) + \bm{g}, {\bm{\beta}}_{k+1} - \bm{z}\rangle +
    \\
    \frac{L}{2}\|{\bm{\beta}}_{k+1} - {\bm{\beta}}_{k}\|^{2}
\end{multline}
Recall that: 
\begin{align*}
    \widehat{\bm{\beta}}_{k+1} & = \text{prox}_{\eta,h}(\bm{\beta}_{k} - \eta \nabla f(\bm{\beta_{k}}))
    \\
    & = \text{argmin}_{\bm{\beta}'} \text{ } h(\bm{\beta}') + \frac{1}{2\eta}\|(\bm{\beta}_{k} - \eta \nabla f(\bm{\beta_{k}})) - \bm{\beta}'\|^{2}
\end{align*}
Imposing optimality conditions, we get:
\begin{align}
    \nonumber
    & \bm{0} \in \partial h(\widehat{\bm{\beta}}_{k+1}) + \nabla f({\bm{\beta}}_{k}) + \frac{1}{\eta} (\widehat{\bm{\beta}}_{k+1} - {\bm{\beta}}_{k})
    \\
    \label{eq-23}
    & \implies -\nabla f({\bm{\beta}}_{k}) + \frac{1}{\eta}({\bm{\beta}}_{k} - \widehat{\bm{\beta}}_{k+1}) \in \partial h(\widehat{\bm{\beta}}_{k+1}) 
\end{align}
Using the above value as $\bm{g}$ in (\ref{eq-22}), we get:
\begin{multline}
    \label{eq-24}
    \phi({\bm{\beta}}_{k+1}) \leq \phi(\bm{z}) + \langle \frac{1}{\eta}({\bm{\beta}}_{k} - \widehat{\bm{\beta}}_{k+1}) , {\bm{\beta}}_{k+1} - \bm{z}\rangle +
    \\
    \frac{L}{2}\|{\bm{\beta}}_{k+1} - {\bm{\beta}}_{k}\|^{2}
\end{multline}
Expressing ${\bm{\beta}}_{k} - \widehat{\bm{\beta}}_{k+1}$ as $({\bm{\beta}}_{k} - {\bm{\beta}}_{k+1}) + ({\bm{\beta}}_{k+1} - \widehat{\bm{\beta}}_{k+1})$, we can rewrite (\ref{eq-24}) as:
\begin{multline}
    \label{eq-25}
    \phi({\bm{\beta}}_{k+1}) \leq \{\phi(\bm{z}) + \frac{1}{\eta} \langle {\bm{\beta}}_{k} - {\bm{\beta}}_{k+1} , {\bm{\beta}}_{k+1} - \bm{z}\rangle +
    \\
    \frac{L}{2}\|{\bm{\beta}}_{k+1} - {\bm{\beta}}_{k}\|^{2}\} + \frac{1}{\eta} \underbrace{\langle {\bm{\beta}}_{k+1} - \widehat{\bm{\beta}}_{k+1}, {\bm{\beta}}_{k+1} - \bm{z}\rangle}_{\text{(I)}}
\end{multline}
Now since ${\bm{\beta}}_{k+1} = \text{P}_\mathcal{S}(\widehat{\bm{\beta}}_{k+1})$, by the optimality condition of projection, we must have (with $\bm{z} \in \mathcal{S}$):
\begin{equation}
    \label{eq-26}
    \langle {\widehat{\bm{\beta}}_{k+1} - \bm{\beta}}_{k+1}, \bm{z} - {\bm{\beta}}_{k+1}\rangle = \text{(I)}\leq 0
\end{equation}
So finally, for all $\bm{z} \in \mathcal{S}$, we must have:
\begin{multline}
    \label{eq-27}
    \phi({\bm{\beta}}_{k+1}) \leq \phi(\bm{z}) + \frac{1}{\eta} \langle {\bm{\beta}}_{k} - {\bm{\beta}}_{k+1} , {\bm{\beta}}_{k+1} - \bm{z}\rangle +
    \\
    \frac{L}{2}\|{\bm{\beta}}_{k+1} - {\bm{\beta}}_{k}\|^{2}
\end{multline}
Firstly, substituting $\bm{z} = \bm{\beta}_{k}$ in (\ref{eq-27}), we get with $\eta < 1/L$:
\begin{align}
    \nonumber
    \phi({\bm{\beta}}_{k+1}) & \leq \phi(\bm{\beta}_{k}) - \frac{1}{\eta} \|{\bm{\beta}}_{k+1} - {\bm{\beta}}_{k}\|^{2} + \frac{L}{2}\|{\bm{\beta}}_{k+1} - {\bm{\beta}}_{k}\|^{2}
    \\
    \label{eq-28}
    & \leq \phi(\bm{\beta}_{k}) - \frac{1}{2\eta} \|{\bm{\beta}}_{k+1} - {\bm{\beta}}_{k}\|^{2}
\end{align}
Thus, we have $\phi({\bm{\beta}}_{k+1}) \leq \phi({\bm{\beta}}_{k})$ with $\eta < 1/L$.

Next, we substitute $\bm{z} = \bm{\beta}^{*}$ in (\ref{eq-27}). From this, we get with $\eta < 1/L$:
\begin{multline*}
    \phi({\bm{\beta}}_{k+1}) \leq \phi(\bm{\beta}^{*}) + \frac{1}{\eta} \langle {\bm{\beta}}_{k} - {\bm{\beta}}_{k+1} , {\bm{\beta}}_{k+1} - \bm{\beta}^{*}\rangle +
    \\
    \frac{1}{2\eta}\|{\bm{\beta}}_{k+1} - {\bm{\beta}}_{k}\|^{2}
\end{multline*}
Manipulating the above equation, we get:
\begin{equation}
    \label{eq-29}
    \phi({\bm{\beta}}_{k+1}) - \phi(\bm{\beta}^{*}) \leq \frac{1}{2\eta}(\|{\bm{\beta}}_{k} - \bm{\beta}^{*}\|^{2} - \|{\bm{\beta}}_{k+1} - \bm{\beta}^{*}\|^{2})
\end{equation}
Summing (\ref{eq-29}) from $k=1,\ldots,T$ and using the fact that $\phi({\bm{\beta}}_{k+1}) \leq \phi({\bm{\beta}}_{k})$ with $\eta < 1/L$ (see (\ref{eq-28})), we get:
\small
\begin{align}
    \nonumber
    T(\phi({\bm{\beta}}_{T+1}) - \phi(\bm{\beta}^{*})) & \leq \frac{1}{2\eta}(\|{\bm{\beta}}_{1} - \bm{\beta}^{*}\|^{2} - \|{\bm{\beta}}_{T+1} - \bm{\beta}^{*}\|^{2})
    \\
    \label{eq-30}
    & \leq \frac{1}{2\eta}\|{\bm{\beta}}_{1} - \bm{\beta}^{*}\|^{2}
\end{align}
\normalsize
Thus, we get:
\begin{equation}
    \label{eq-31}
    \phi({\bm{\beta}}_{T+1}) - \phi(\bm{\beta}^{*}) \leq \frac{1}{2\eta T}\|{\bm{\beta}}_{1} - \bm{\beta}^{*}\|^{2}
\end{equation}

Recall that $\bm{\beta}_{1} = \Big[\frac{1}{k_{1}^{1/2p}}\mathbf{1}_{k_{1}},\frac{1}{k_{2}^{1/2p}}\mathbf{1}_{k_{2}},\ldots,\frac{1}{k_{n}^{1/2p}}\mathbf{1}_{k_{n}}\Big]^{T}$. 
Since $\bm{\beta}^{*} \in \mathcal{S}$, we have:
\begin{align}
    \nonumber
    \|{\bm{\beta}}_{1} - \bm{\beta}^{*}\|^{2} & \leq \sum_{i=1}^{n}\sum_{j=1}^{k_{i}}\Bigg(-1-\frac{1}{k_{i}^{1/2p}}\Bigg)^{2}
    \\
    \label{eq-32}
    & = \sum_{i=1}^{n} k_{i}\Bigg(1+\frac{1}{k_{i}^{1/2p}}\Bigg)^{2}
\end{align}
Substituting (\ref{eq-32}) in (\ref{eq-31}), we get:
\begin{align}
    \label{eq-33}
    \phi({\bm{\beta}}_{T+1}) - \phi(\bm{\beta}^{*}) & \leq \frac{1}{2 \eta T} \sum_{i=1}^{n} k_{i}\Bigg(1+\frac{1}{k_{i}^{1/2p}}\Bigg)^{2}
    \\
    \label{eq-34}
    & \leq \frac{1}{2 \eta T} \sum_{i=1}^{n} k_{i} . 2^2 = \frac{2}{\eta T}\sum_{i=1}^{n} k_{i}
\end{align}
(\ref{eq-33}) proves the first result of the theorem. We use (\ref{eq-34}) to show the second result of the theorem.
\begin{equation*}
    \frac{2}{\eta T}\sum_{i=1}^{n} k_{i} = \epsilon \implies \phi({\bm{\beta}}_{T+1}) - \phi(\bm{\beta}^{*}) \leq \epsilon
\end{equation*}
But:
\begin{equation*}
    \label{eq-35}
    \frac{2}{\eta T}\sum_{i=1}^{n} k_{i} = \epsilon \implies T = \frac{2}{\eta \epsilon}\sum_{i=1}^{n} k_{i}
\end{equation*}
Thus we need $T = \mathcal{O}(\sum_{i=1}^{n} k_{i}/\epsilon)$ iterations to achieve $\phi({\bm{\beta}}_{T+1}) - \phi(\bm{\beta}^{*}) \leq \epsilon$.
\end{proof}

\section{Preliminary Empirical Validation}
During implementation, the adversarial text generation would proceed in two steps:\\
(A) \textit{Candidate Set Generation}: This subroutine would generate potential substitution candidates for each token i.e. in the context of our formulation in Section ~\ref{sec:intro}, for a given set $\sS = \{e_1, e_2 , \dots , e_n\} \in \mathcal{X}^n$ this subroutine would generate for each feature $e_i \in \mathcal{X}, \; k_i -1$ possible replacements denoted by $e_i^j, \; \forall j \in [k_i - 1]$ with $e_i^0 = e_i, \; \forall e_i \in \mathcal{X}$. (B)\textit{Combinatorial Search}: Now we solve Equation \eqref{eq:adv_attack} by some combinatorial approach. For example in this paper we discussed Algorithm \ref{algo:greedy} and \ref{alg:ppgd} for the second sub-routine.  
\subsection{Candidate Set Generation}
The first step in adversarial text generation is to find suitable replacement candidates for each token in the text. This problem can be viewed in several ways. One approach would be to think of this as a synonym finding problem where given a token the objective is to find semantically equivalent tokens.
The first natural approach would be to project the token into the continuous embedding space like Glove \cite{pennington2014glove} and then use similarity measure like cosine similarity or euclidean distance to find nearest neighbors in the embedding space. However, this doesn't quite work as well as one would hope for. Pre-trained embeddings like Glove are learned using the co-occurance information of tokens using CBOW or skip-gram based approaches \cite{mikolov2013efficient, goldberg2014word2vec}. Co-occurance information can be misleading when we are looking for a particular semantic relationship like synonym or antonym. To motivate: let's assume good and bad which are antonyms i.e. have exactly the opposite meaning. However, in the Glove embedding space they will be extremely close to each other , which makes sense since they appear in a similar contexts. Another popular approach would have been to use a pre-trained language model like BERT \cite{devlin2018bert}, then given a text mask the token we need replacement for and let the model predict the token. However, this kind of approach also suffers from a similar shortcomings as Word2Vec training approach. \\
Motivated by success of \cite{mrkvsic2016counter, wieting2015paraphrase} on word semantic equivalence tasks like \cite{hill2015simlex,finkelstein2002placing} we take a similar approach to generate our candidate set. 

Similar to the approach by \cite{mrkvsic2016counter} we start from Glove embeddings and then re-align the word vectors under three constraints: synonyms are placed close to each other(S), antonyms are pushed further away from each other(A) while making sure that the topology of the vector space is preserved (VS). More specifically we start from embedding vector space $V=\left\{\mathbf{v}_{1}, \mathbf{v}_{2}, \dots, \mathbf{v}_{N}\right\} $ and inject semantic relationship to learn a new representation $V=\left\{\mathbf{v'}_{1}, \mathbf{v'}_{2}, \dots, \mathbf{v'}_{N}\right\}$. Now we inject the following constraints for semantic equivalence: 
\begin{equation}
    \text{A =} \sum_{(u, w) \in A} \tau\left(\delta-d\left(\mathbf{v}_{u}^{\prime}, \mathbf{v}_{w}^{\prime}\right)\right)
\end{equation}
\begin{equation}
    \text{S =} \sum_{(u, w) \in S} \tau\left(d\left(\mathbf{v}_{u}^{\prime}, \mathbf{v}_{w}^{\prime}\right)-\gamma\right)
\end{equation}
\begin{equation}
    \text{VS = }\sum_{i=1}^{N} \sum_{j \in N(i)} \tau\left(d\left(\mathbf{v}_{i}^{\prime}, \mathbf{v}_{j}^{\prime}\right)-d\left(\mathbf{v}_{i}, \mathbf{v}_{j}\right)\right)
\end{equation}
The problem of finding the semantically modified vector space reduces to the following optimization problem: 
\begin{equation}
    C\left(V, V^{\prime}\right)=k_{1} \mathrm{A}\left(V^{\prime}\right)+k_{2} \mathrm{S}\left(V^{\prime}\right)+k_{3} \mathrm{VS}\left(V, V^{\prime}\right)
\end{equation}
Our candidate selection subroutine takes in a token , projects the token into this modified vector space and then runs a k nearest neighbor style neighbor search based on similarity score (euclidean distance) with a pre-defined k. 
We further constraint that any selected candidate should be within an euclidean ball of radius $\epsilon$ of the input token. Formally for two tokens to co-exist in our candidate set (\ref{knn1}) needs to hold for a pre-specified threshold $\eta$ where $R$ is the euclidean distance of the furthest token in the vocabulary.: 
\begin{equation}
\label{knn1}
    \begin{array}{l}
        ||V'_{i} - V'_{j}||_{2} \leq \epsilon \\
        \ni \epsilon = \eta \times R
    \end{array}
\end{equation}

\subsection{Data Sets}
\label{sec:dataset}
We show the effectiveness of the proposed Continuous Relaxation approach on three different tasks as described below: 

\paragraph{\textbf{Sentiment Classification}}
The task is to look at a piece of text and assign a positive or negative label to the task. For this task we chose the popular IMDB MR data-set \cite{maas2011learning}. It is a binary classification data-set with 25k positive and negative reviews.

\paragraph{\textbf{News Classification}} Given a piece of news snippet the task is to classify it into one of the categories. Here, we will use AG News corpus \cite{zhang2015character} which is a relatively large dataset with 30,000 training and 1,900 testing examples for each of the fours classes.

\paragraph{\textbf{Question Classification}} 
To see the effectiveness of our algorithm in context of relatively short text we chose the TREC-QC dataset \cite{li2002learning} where the task is to classify the question (usually a single sentence) into one of the 6 categories. This dataset has about 5500 training and about 500 test examples. 

\subsection{Greedy Baseline}
\label{greedy}
We compared with the greedy approach used extensively in the existing literature for a similar setting \cite{lei2018discrete, alzantot2018generating, kuleshov2018adversarial} essentially using the Algorithm \ref{algo:greedy} with the uniform matroid constraint. 
The approach is quite simple - you start from the first token the the input text, greedily choose the optimal candidate that maximizes the adversarial loss, fix that token and continue to the next token until end of the text is reached. 
This baseline is more aggressive and is unconstrained in semantic coherent space, as argued in \cite{lei2018discrete, alzantot2018generating} constraining the search space to semantically coherent space i.e. using a language model perplexity constraint would improve the semantic coherence but obviously lead to worse performance in terms of adversarial accuracy. However, we use this as a baseline for comparison in terms of attack success (in terms ) since this loosely speaking is an upper bound on the performance of all the methods that use greedy approach for the candidate selection problem. It will be interesting to see a comparison of the perplexity scores of different algorithms to compare semantic equivalence for a fairer comparison. And intuitively, our algorithm is currently under-stated as it is in theory always going to produce semantically more coherent text while achieving high accuracy. 
\begin{figure*}
    \centering
    \includegraphics[width=15cm]{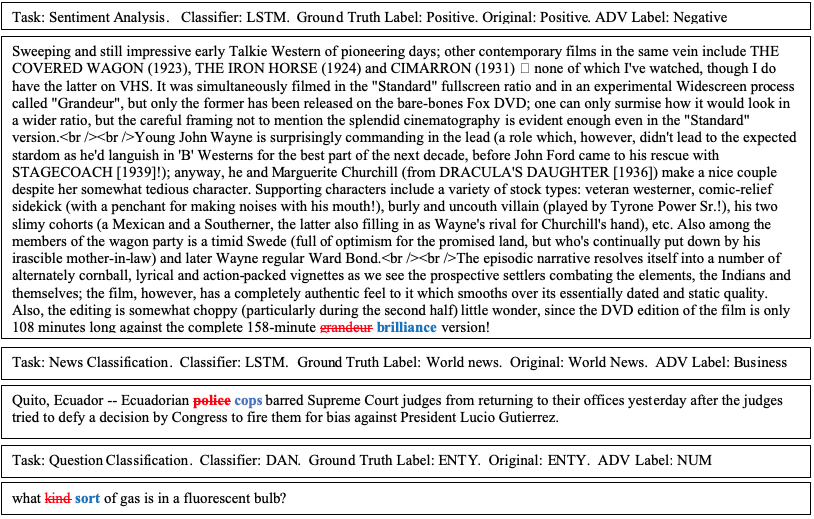}
    \caption{Examples of generated adversarial examples. We show three examples here from three tasks, and two models The color red denote the original word. The color blue denote the adversarially perturbed word. A more detailed result can be found in [\ref{exp}]}.
    \label{fig:text_example}
\end{figure*}
\begin{table}[!htb]
\centering
\begin{tabular}{@{}llll@{}}
\toprule
\multicolumn{4}{c}{\textbf{Continuous Relaxation Approach (Algorithm \ref{alg:ppgd})}}\\ 
\midrule
Dataset & \begin{tabular}[c]{@{}l@{}}Original Acc.(\%)\end{tabular} & \begin{tabular}[c]{@{}l@{}}Adversarial Acc.(\%)\end{tabular} & \begin{tabular}[c]{@{}l@{}}Perturbation (\%)\end{tabular} \\ \cmidrule(r){1-4}
IMDB & 87  & 18 & 4.6 \\
AG News & 88.75 & 26.5 & 16.5\\
\bottomrule
\midrule
\multicolumn{4}{c}{\textbf{Greedy Approach (Algorithm \ref{algo:greedy})}}\\ 
\midrule
IMDB    & 87 & 62  & 10 \\
AG News & 88.75 & 32.5 & 20 \\
\bottomrule
\end{tabular}
\caption{
Comparison of Greedy \cite{lei2018discrete} (Algorithm \ref{algo:greedy}) with the proposed relaxed Algorithm \ref{alg:ppgd} on different tasks. 50 samples were chosen randomly from the test set to attack and evaluate the attack models.  
We present the accuracy of the model before attack on this randomly chosen subset (original acc). The ADV acc column shows the accuracy of the model when those same examples were adversarially perturbed using Algorithm \ref{alg:ppgd}. The last column reports the amount of adversarial perturbation (\% of tokens modified) in the entire set.}
\label{tab:DISCO}
\end{table}
\subsection{Experimental Setup}
For the experiments have used the extremely popular and successful recurrent model LSTM \cite{hochreiter1997long} with hidden size 200 with network dropout \cite{srivastava2014dropout} rate of 0.3. Our networks are also trained with batch normalization \cite{ioffe2015batch} to reduce internal co-variance and ensure faster convergence. The models were trained using Adam \cite{kingma2014adam}.
While running Algorithm \ref{alg:ppgd} - in all experiments, we started with $\beta_{i}^{(1)} = 10$ and $\beta_{i}^{(j)} = 0.05$ for $j > 1$, $\forall i \in [n]$. We used the following values of $\lambda$, $\gamma$ and $\kappa$:
\[\lambda = 3C(s)/(10n + 0.05\sum_{i=1}^{n}(k_{i}-1))\] 
\[\gamma = 6C(s)/\sqrt{n}\]
\[\kappa = 0.2\]
To solve our objective function, we used Adam with initial learning rate of 1 and a maximum epoch budget of 100.

\subsection{Results}
\label{result}
Table ~\ref{tab:DISCO}shows our results on the different datasets to adversarially attack LSTM model. We also present some sample adversarial text produced by Algorithm \ref{alg:ppgd} in Fig (\ref{fig:text_example}).
To compare Greedy Baseline and Continuous Approximation, we randomly sample $n = 50$ samples from each class and call this our attack set. On this attack set the attack algorithms are evaluated using the following metrics:

\paragraph{\textbf{Initial Accuracy}} Initial Accuracy of the model before the attack. A higher number indicate a well trained model.

\paragraph{\textbf{Adversarial Accuracy}}
Simple accuracy after the attack i.e. number of correct predictions to the total number of adversarially perturbed test samples. A lower number implies a more successful attack.

\paragraph{\textbf{Perturbation}}
While accuracy is a good way to measure the success of an adversarial attack algorithm, it is also essential to ensure amount of perturbation used to fool the model. In theory, one can arbitrarily change the input sentence for example paraphrase based methods \cite{iyyer2018adversarial} can arbitrarily change the syntactic structure of the text. However, by definition adversarial examples are to be impermeable to human judge and arbitrarily changing the syntactic structure of the text would violate this. To capture this, we compute Perturbation which is defined as the Average value of $\frac{\text{No. of tokens changed}}{\text{Total no. of tokens}}$ per example.
Note that, a lower number is better indicating the adversarial
model is able to perform well with lesser syntactic perturbation. 

\paragraph{}
In summary, we observed that: Algorithm \ref{alg:ppgd} seems to outperform the greedy approach in all tasks in terms of accuracy which is not so surprising. In principle since our baseline is supposed to be an upper bound in terms of performance among all the variants of greedy algorithm implying our algorithm might be producing state-of-the-art results. While, achieving higher success rate , it seems to be able to do so with much less perturbation as can be seen in Table~\ref{tab:DISCO}. 
Additionally, as can be seen subjectively in Fig~\ref{fig:text_example} our approach is also able to produce high quality adversarial text. 
\label{exp}
\bibliographystyle{alpha}
\bibliography{bibs/comb_opt_ref,bibs/decoding_ref}

\section*{Acknowledgements}
This work is supported in part by NSF grants CCF-1564000, IIS-1546452 and HDR-1934932.
\end{document}